\def\year{2019}\relax
\documentclass[letterpaper]{article} %DO NOT CHANGE THIS
\usepackage{aaai19}  %Required
\usepackage{times}  %Required
\usepackage{helvet}  %Required
\usepackage{courier}  %Required
\usepackage{url}  %Required
\usepackage{graphicx}  %Required
\usepackage{enumerate}
\usepackage{amsmath}

\begin{document}
\title{Chi-Square Test Neural Network: A New Binary Classifier based on Backpropagation Neural Network}
\author{Yuan Wu , Lingling Li \and Lian Li
\\
School of Information Science and Engineering\\
 Lanzhou University, Lanzhou 730000, China.\\
}

\maketitle
\begin{abstract}
We introduce the chi-square test neural network ($\chi^2 NN$): a single hidden layer backpropagation neural network using chi-square test theorem to redefine the cost function and the error function. The weights and thresholds are modified using standard backpropagation algorithm. The proposed approach has the advantage of making consistent data distribution over training and testing sets. It can be used for binary classification. The experimental results on real world data sets indicate that the proposed algorithm can significantly improve the classification accuracy comparing to related approaches.
\end{abstract}

\section{Introduction}
\noindent Artificial neural networks (ANNs) have the abilities of mimicking complex and non-linear relationships by using many non-linear processing units called neurons. It has advantages of strong adaptability, flexible modeling capability and parallel computing abilities \cite{wu2017nondestructive}. ANN presents a parameterized, non-linear mapping between inputs and outputs \cite{irani2011evolving}. The relationship between neurons can be learnt through training based on the features presented by the data \cite{lin2008application}. This data-driven approach can be used to tackle many different problems, such as classifying nonlinearly separable patterns and approximating arbitrarily continuous functions. ANN is one of the most commonly used form of supervised learning algorithms. Meanwhile, the backpropagation neural network (BPNN) is the most commonly used ANNs. BPNN uses the back propagation-learning algorithm, which is a mentor-learning algorithm of gradient descent \cite{zhang1998forecasting}. According to the theory, BPNN has the properties of forward propagation of signals and back propagation of errors. The learning algorithm tunes the weights and thresholds in BPNN automatically in order to minimize the error so that a single hidden layer BPNN can generally approximate any nonlinear function with arbitrary precision \cite{aslanargun2007comparison}.

The general structure of BPNN consists of three layers: an input layer, a hidden layer and an output layer (Fig.1). The BPNN model formulation includes 4 steps:
\begin{enumerate}[step 1]
\item Initializing the weights and thresholds in the BPNN model randomly;
\item transmitting the information from the input layer to the output layer, and obtaining the output values;
\item Calculating the mean square error (MSE) between the output values and the actual values;
\item If the MSE achieves the goal setting, the weights and thresholds are determined, so the training process of the model can be finished; otherwise, adjusting the weights and thresholds through gradient descent and then going to Step2.
\end{enumerate}

\begin{figure}
  \includegraphics[clip=ture,width=0.5\textwidth]{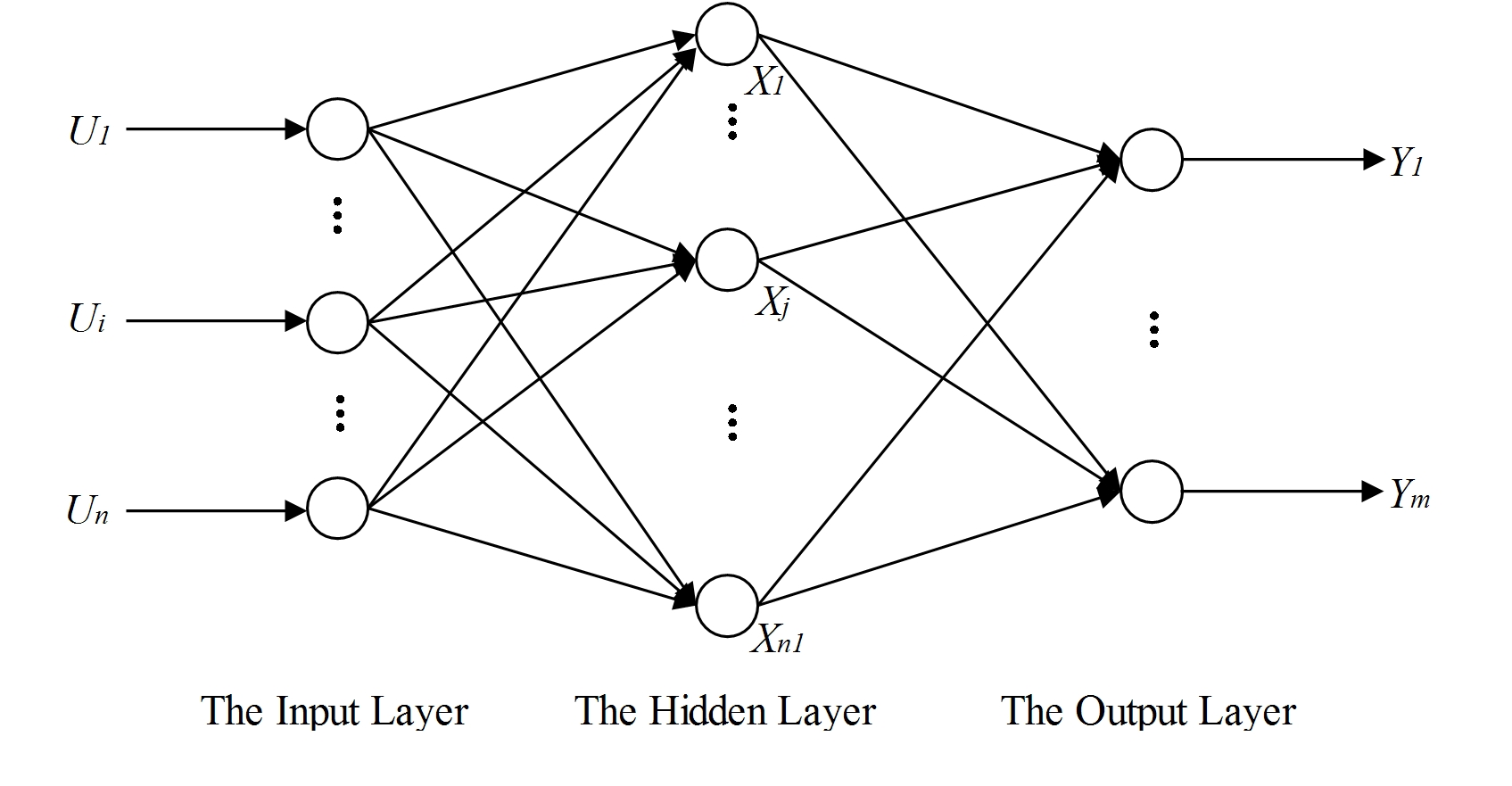}
\caption{The general structure of BPNN.}
\label{fig:1}       % Give a unique label
\end{figure}

However, BPNN has some inherent problems when facing non-linear classification problems, such as it can easily fall into local minimum point rather than global minimum point and its convergent speed is very slow \cite{gori1992problem}. An alternative to BPNN that has been used in classification is the probabilistic neural network (PNN) \cite{specht1990probabilistic}. In this paper, a novel $\chi^2 NN$ which can be used for binary classification was proposed. The experimental results on real world data sets demonstrated that the proposed model can significantly outperform other traditional classifiers.

\section{Model Description}

The binary classification problem can be defined as follows. Given n training data ${(x_1,y_1 ),(x_2,y_2 ),¡­,(x_n,y_n)}$, in which $x_i\in R^n$ represents the input including n features and $y_i\in \{0,1\}$ represents the class. So we need to establish an classifier $f:R^n\rightarrow \{0,1\}$ that maximizes the probability that $f(x_i )=y_i,(i=1,2,¡­,n)$. For traditional BPNN model, there exists a real valued function: $f(x,w,b)=<w,x>+b$, in which $w=(w_1,w_2,¡­w_n)\in R^n$ and $b=(b_1,b_2,¡­b_n)\in R^n$. Here $<w,x>$ denotes the dot product of two vectors $w$ and $x$. Then a reference value $\alpha $ should be set, if the result $\hat{y}_i>\alpha$, $\hat{y}_i$ belongs to $\{1\}$; otherwise $\hat{y}_i$ belongs to $\{0\}$.

Suppose that N observations in a random sample from a population are classified into M mutually exclusive sections with respective observed numbers $v_i,(i=1,2,¡­M)$, and a null hypothesis gives the probability $p_i$ that an observation falls into the i-th section. So we have the expected numbers $m_i=Np_i,(i=1,2,¡­,M)$, where

\begin{equation}
  \sum_{i=1}^{M}m_i=N\sum_{i=1}^{M}p_i=\sum_{i=1}^{M}v_i
\label{eq:2}
\end{equation}

According to \cite{pearson1900x}, there exists:
\begin{equation}
  \eta =\sum_{i=1}^{M}\frac{(v_i-m_i)^2}{m_i}
\label{eq:3}
\end{equation}

If the expected numbers $m_i$ are large enough and the observed numbers $v_i$ are normally distributed, $\eta$ follows the chi-square distribution with $M-1$ degrees of freedom.
In $\chi^2 NN$(Fig.2), $\eta$ is used as the cost function, the initialization consists of 7 steps:

\begin{figure}
  \includegraphics[clip=ture,width=0.5\textwidth]{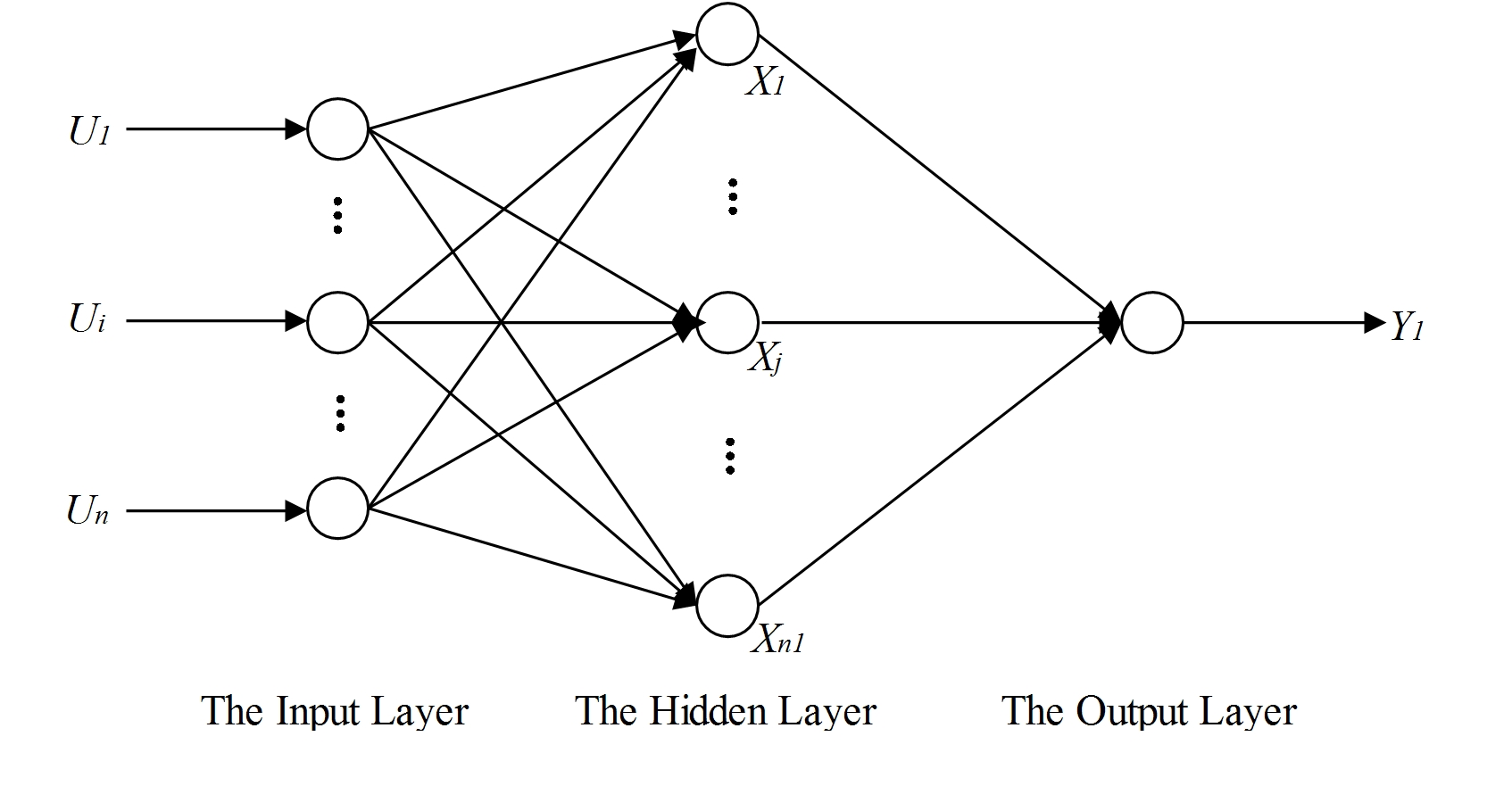}
\caption{The structure of $\chi^2 NN$.}
\label{fig:2}       % Give a unique label
\end{figure}

\begin{enumerate}[step 1]
\item Extracting main features from raw data using PCA because neural networks may fail with highly dimensional input, and using these features as the input vector.
\item For every dimension of the processed data, it can be divided into K mutually exclusive sections with equal length, so we have $K^L$ sections (L is the dimension number) in the input space. Here we have $M=K^L$.
\item Dividing the data set into two parts: the training set and the testing set, the samples in both sets are selected randomly, so it can guarantee that the training set and the testing set are consistent with the same data distribution \cite{lindgren2017statistical}.
\item Using the training set to calculate the numbers $N_i$  and $p_i$(i=1,¡­,M), $N_i$ represents the number of data point which falls into the i-th section, $\sum_{i=1}^{M}N_i=N$.
\item Setting the activation function for the hidden layer:

\begin{equation}
  f(x)=sigmoid(x)=\frac{1}{1+e^{-x}}
\label{eq:4}
\end{equation}

and that for the output layer:
\begin{equation}
  g(x,y)=\left\{
\begin{aligned}
1&,&x>0.5 \\
0&,&x\leq0.5
\end{aligned}
\right.
\label{eq:5}
\end{equation}

\item Defining $v_i$ as follows:

\begin{equation}
  v_i=\sum_{l=1}^{N_i}O_l
\label{eq:6}
\end{equation}

\begin{equation}
  O_l=g(\sum_{j=1}^{m}\omega _jO_{lj}+\theta )
\label{eq:7}
\end{equation}

where $O_l$ represents the l-th output obtained from the neuron in the output layer, $m$ indicates the number of neurons in the hidden layer, $O_{lj}$ represents the l-th output obtained from the j-th neuron in the hidden layer, $w_j$ represents the weight which connects the j-th neuron in the hidden layer and the neuron in the output layer, $\theta$ represents the bias for the neuron in the output layer.

\item Reformulating the error function as follows:
\begin{equation}
  E=\frac{1}{2}\sum_{i=1}^{m}(\frac{v_i}{N}-p_i)^2
\label{eq:8}
\end{equation}
where the error function E has the same monotonicity with the cost function $\eta$.

\item Assigning the weights and thresholds in the network with random values.
\end{enumerate}

\section{Modification of the parameters}
In traditional BPNNs, the weights and thresholds are automatically adjusted using the gradient descent, the modification to these parameters is aimed at achieving the minimum MSE values between the output values and the actual values. According to the back propagation algorithm, the modification to the weights and thresholds in the network should be done along the negative gradient direction, so we have
\begin{equation}
  v\leftarrow v+\Delta v
\label{eq:9}
\end{equation}

\begin{equation}
  \Delta v=-\rho \frac{\partial E }{\partial v}
\label{eq:10}
\end{equation}

where $v$ represents a parameter in the network and $\rho (\rho\in(0,1))$ represents the learning rate.
In $\chi^2NNs$, we still use the gradient descent algorithm to modify the parameters in networks. The $\chi^2NN$ is running on the assumption that when data sets are consistent to the same data distribution, the proportion of the samples which belong to one class in each section of the input space for one data set should be equal to that for the other data sets. The training set is used for adjustment of the parameters in $\chi^2NN$.

For the parameters between the hidden layer and the output layer, there exists:
\begin{equation}
  \frac{\partial E}{\partial \omega _j}=\sum_{i=1}^{M}\frac{\partial E}{\partial v _i}*\frac{\partial v_i}{\partial \omega_j}=\sum_{i=1}^{M}\frac{1}{N}(\frac{v_i}{N}-p_i)*\frac{\partial v_i}{\partial \omega _j}
\label{eq:11}
\end{equation}

\begin{equation}
\frac{\partial E}{\partial \theta }=\sum_{i=1}^{M}\frac{\partial E}{\partial v _i}*\frac{\partial v_i}{\partial \theta }=\sum_{i=1}^{M}\frac{1}{N}(\frac{v_i}{N}-p_i)*\frac{\partial v_i}{\partial \theta }
\label{eq:12}
\end{equation}

in which
\begin{equation}
  \frac{\partial v_i}{\partial \omega _j}=\sum_{l=1}^{N_i}\frac{\partial v_i}{\partial O_l}*\frac{\partial O_l }{\partial \omega _j}
\label{eq:13}
\end{equation}

\begin{equation}
  \frac{\partial v_i}{\partial \theta }=\sum_{l=1}^{N_i}\frac{\partial v_i}{\partial O_l}*\frac{\partial O_l }{\partial \theta }
\label{eq:14}
\end{equation}

since $g(x)$ is the transition function and $g(x)$ is a not monotone decreasing function, ${g}'(x)$ won't affect the direction that the gradient descends, we set ${g}'(x)=\xi$, $\xi$ is a constant. So we have

\begin{equation}
  \frac{\partial E}{\partial \omega _j}=\sum_{i=1}^{M}[ \frac{\xi}{N}(\frac{v_i}{N}-p_i)*\sum_{l=1}^{N_i}O_{lj}]
\label{eq:15}
\end{equation}

\begin{equation}
  \frac{\partial E}{\partial \theta }=\sum_{i=1}^{M}[ \frac{1}{N}(\frac{v_i}{N}-p_i)*\sum_{l=1}^{N_i}\xi ]
\label{eq:16}
\end{equation}

\begin{equation}
  \omega _j=\omega _j-\rho*\sum_{i=1}^{M}[ \frac{\xi}{N}(\frac{v_i}{N}-p_i)*\sum_{l=1}^{N_i}O_{lj}]
\label{eq:17}
\end{equation}

\begin{equation}
  \theta=\theta-\rho*\sum_{i=1}^{M}[ \frac{1}{N}(\frac{v_i}{N}-p_i)*\sum_{l=1}^{N_i}\xi ]
\label{eq:18}
\end{equation}

For the parameters between the input layer and the output layer, since
\begin{equation}
  O_{lj}=f(I_{lj})=sigmoid(I_{lj})
\label{eq:19}
\end{equation}

\begin{equation}
  I_{lj}=\sum_{k=1}^{r}\omega _{kj}I_{lk}+\theta _j
\label{eq:20}
\end{equation}

in which $I_{lj}$ represents the l-th input for the j-th neuron in the hidden layer, $r$ is the number of neurons in the input layer, $w_{kj}$ represents the weight connects the k-th neuron in the input layer and the j-th neuron in the hidden layer, $I_{lk}$ represents the l-th input for the k-th neuron in the input layer and $\theta_j$ represents the bias for the j-th neuron in the hidden layer. So there exists:

\begin{equation}
\begin{aligned}
&\frac{\partial E}{\partial \omega _{kj}}=\sum_{i=1}^{M}\frac{\partial E}{\partial v_i}*\frac{\partial v_i}{\partial O_l}*\frac{\partial O_l}{\partial O_{lj}}*\frac{\partial O_{lj}}{\partial I_{lj}}*\frac{\partial I_{lj}}{\partial \omega _{kj}}\\
&=\sum_{i=1}^{M}[\frac{1}{N}(\frac{v_i}{N}-p_i)*\sum_{l=1}^{N_i}\xi *\omega _{kj}*O_{lj}*(1-O_{lj})*I_{lk}]
\end{aligned}
\label{eq:21}
\end{equation}

\begin{equation}
\begin{aligned}
  &\frac{\partial E}{\partial \theta _j}=\sum_{i=1}^{M}\frac{\partial E}{\partial v_i}*\frac{\partial v_i}{\partial O_l}*\frac{\partial O_l}{\partial O_{lj}}*\frac{\partial O_{lj}}{\partial I_{lj}}*\frac{\partial I_{lj}}{\partial \theta _j}\\
&=\sum_{i=1}^{M}[\frac{1}{N}(\frac{v_i}{N}-p_i)*\sum_{l=1}^{N_i}\xi *\omega _{kj}*O_{lj}*(1-O_{lj})]
\end{aligned}
\label{eq:22}
\end{equation}

\begin{equation}
  \omega _{kj}=\omega _{kj}-\rho*\sum_{i=1}^{M}[\frac{1}{N}(\frac{v_i}{N}-p_i)*\sum_{l=1}^{N_i}\xi *\omega _{kj}*O_{lj}*(1-O_{lj})*I_{lk}]
\label{eq:23}
\end{equation}

\begin{equation}
  \theta _j=\theta _j-\rho*\sum_{i=1}^{M}[\frac{1}{N}(\frac{v_i}{N}-p_i)*\sum_{l=1}^{N_i}\xi *\omega _{kj}*O_{lj}*(1-O_{lj})]
\label{eq:24}
\end{equation}

Since the cost function $\eta $ follows the chi-square distribution with $M-1$ degrees of freedom, we set $\varepsilon =\chi^2 (M-1)$, if $\eta<\varepsilon$, the weights and thresholds are determined, and the model construction is finished; otherwise, the iteration continues.

\section{Experiments}
We conducted experiments on several publicly available data sets: Iris, India Liver Patient Dataset (ILPD), Banknote Authentication (BA), Breast Cancer Wisconsin (BCW) and Balloons. All data sets are from the UCI Machine Learning Repository \cite{asuncion2007uci}. A brief introduction about these data sets are given in Table.1, Table.1 shows the number of features m for the corresponding set and numbers of examples in negative $n_0$ and positive $n_1$ classes respectively. In the experiments below, we compared the performance of the proposed $\chi^2NN$ algorithm with conventional BPNN. In order to compare the performance, the number of hidden neurons in BPNN was the same with that in $\chi^2NN$, and the activation functions used in BPNN are Sigmoid function (for the hidden layer) and Purelin function (for the output layer). Although Support Vector Machine (SVM) is obviously different from ANN, and it was not the objective of this paper to systematically compare the difference between SVM and $\chi^2NN$, the performance comparison between SVM and $\chi^2NN$ was also simply conducted.

\begin{table}[!htb]
\caption{A brief introduction about data sets.}
\label{tab:1}       % Give a unique label
\begin{tabular}{llll}
\hline\noalign{\smallskip}
Data sets & m & $n_0$ & $n_1$ \\
\noalign{\smallskip}\hline\noalign{\smallskip}

Iris &	4 &	50 &50\\
ILPD &	10 &	414 &165\\
BA &	4 &	762 &610\\
BCW&    10 & 444 &239\\
Balloons&4& 41  &35\\
\noalign{\smallskip}\hline
\end{tabular}
\end{table}

It should be noted that the data set Iris consists of 3 different varieties of iris: Setosa, Versicolour and Virginica, each has 50 samples. For using this data set in binary classification, samples of the class Setosa were supposed to be from the first class, that of the class Versicolour were used as the second class. Other data sets initially consist of 2 classes.

In our experiments, the $\chi^2NN$ binary classifier was established by using 5 steps:
\begin{enumerate}[step 1]
\item Using PCA to extract main features from the raw data based on the criteria that the selected PCs¡¯ accumulative contribute rate should be no less than 90\%.
\item Setting the parameters $K$, $\xi$, $\rho$ and the number of neurons in the hidden layer to be 2, 0.5, 0.1 and 10, respectively.
\item All samples from the data set were randomly divided into the training set and testing set, the training set contained 90\% samples.
\item The weights and thresholds in $\chi^2NN$ were tuned automatically on the basis of training set.
\item The classification accuracy for every decision strategy is determined based on the testing set in accordance with every strategy.
\end{enumerate}

The classification accuracies are computed as average values by means of the random selection of training and testing sets from data sets 20 times.

Table.2 lists the accumulative contribution rates of the first 5 Principle Components (PCs) for each data set. We used 2 PC (Iris), 2 PCs (ILPD), 3 PCs (BA), 5 PCs (BCW) and 4 PCs (Balloons) to construct $\chi^2NN$ classifiers. For SVM classifiers, we used Radial Basis Function (RBF) as the kernel function, the cost parameter C and gamma parameter g were selected by using 10-fold cross-validations. Table.3 shows the optimal values of C and g.

\begin{table}[!htb]
\caption{The accumulative contribution rates of the first 5 PCs for each data set.}
\label{tab:2}       % Give a unique label
\begin{tabular}{llllll}
\hline\noalign{\smallskip}
Data sets & PC1 & PC2 & PC3 & PC4 & PC5 \\
\noalign{\smallskip}\hline\noalign{\smallskip}

Iris &	86.05\%	&96.88\%	&99.42\%	&100\% &N/A\\
ILPD &	62.68\%	&94.34\%	&99.83\%	&99.97\% &100\%\\
BA &	55.39\%	&87.23\%	&95.5\%	    &100\%  &N/A\\
BCW&    69.05\%	&76.25\%	&82.3\%	    &86.74\% &90.64\%\\
Balloons&27.67\%&53.88\%	&77.6\%	    &100\%  &N/A\\
\noalign{\smallskip}\hline
\end{tabular}
\end{table}

\begin{table}[!htb]
\caption{The optimal values of cost parameter and gamma parameter for SVM classifiers.}
\label{tab:3}       % Give a unique label
\begin{tabular}{lll}
\hline\noalign{\smallskip}
Data sets & cost value & gamma valuw \\
\noalign{\smallskip}\hline\noalign{\smallskip}

Iris &	0.04 &	1\\
ILPD &	48.5 &	84.45\\
BA &	1.74 &	256\\
BCW&    84.45 & 0.02\\
Balloons&0.19&1\\
\noalign{\smallskip}\hline
\end{tabular}
\end{table}

\begin{table}[!htb]
\caption{Experimental results in terms of classification accuracy (\%).}
\label{tab:4}       % Give a unique label
\begin{tabular}{llll}
\hline\noalign{\smallskip}
Data sets & $\chi^2NN$ & BPNN & SVM \\
\noalign{\smallskip}\hline\noalign{\smallskip}

Iris &	100 &	100 &100\\
ILPD &	68.97 &	65.52 &68.97\\
BA &	84.67 &	83.21 &86.86\\
BCW&    97.06 & 94.12 &98.53\\
Balloons&87.5& 75  &75\\
\noalign{\smallskip}\hline
\end{tabular}
\end{table}

It is interesting to note that the proposed $\chi^2NN$ outperformed the standard SVM for the Balloons data set and achieved the same classification accuracies with SVM for the Iris and ILPD data sets. At the same time, the $\chi^2NN$ clearly outperformed the conventional BPNN on four of the total five binary data sets. On Iris data set, all the three algorithms gave 100\% classification accuracies. The improvements obtained by the proposed $\chi^2NN$ over the BPNN are largely significant. The experimental results on real world data sets demonstrated that the effectiveness of the proposed $\chi^2NN$ model.

\section{Conclusion}
In this paper, we proposed a classifier for single-hidden layer backpropagation neural networks called chi-square test neural network ($\chi^2NN$). We first used chi-square test theorem to reformulate the cost function and error function for the network. Then we modified the parameter adjustments and the iteration stopping conditions according to the new reformulated error function and cost function. The proposed approach can be used for binary classifications. Moreover, the proposed $\chi^2NN$ can make consistent data distribution over training and testing samples. The experimental results on real world data sets indicated that the proposed algorithm can significantly outperform the traditional BPNN on binary classification tasks.

\bibliographystyle{aaai}
\bibliography{chisquare}

\end{document}